\documentclass[twoside]{article}
\usepackage{fancyhdr}

\usepackage{bm}  
\usepackage{amsmath}
\usepackage{amssymb}
\usepackage{amsfonts}
\usepackage{booktabs}
\usepackage{url}

\usepackage{graphicx}
\usepackage{subfigure} 
\usepackage{comment}

\usepackage{cite} 
\usepackage{float}
\usepackage{lipsum}
\usepackage{color}
\usepackage{macro}

\usepackage{natbib}

\usepackage{algorithm}
\usepackage{algorithmic}
\usepackage{xcolor}

\usepackage{hyperref}

\definecolor{dark-red}{rgb}{0.4,0.15,0.15}
\definecolor{dark-blue}{rgb}{0.15,0.15,0.4}
\definecolor{medium-blue}{rgb}{0,0,0.5}
\hypersetup{
    colorlinks, linkcolor={dark-blue},
    citecolor={dark-blue}, urlcolor={medium-blue}
}

\title{High-Dimensional Probability Estimation with Deep Density Models}
\author{\begin{tabular}{ccc}
    {Oren Rippel\footnote{\url{http://math.mit.edu/\~rippel}}} \qquad &\qquad& Ryan Prescott Adams\footnote{\url{http://people.seas.harvard.edu/\~rpa}} \\
     Massachusetts Institute of Technology,  &\qquad& Harvard University  \\ 
     Harvard University & & \\
   {\tt rippel@math.mit.edu} &\qquad& {\tt rpa@seas.harvard.edu}
  \end{tabular}
}

\begin{document} 

\date{}

\maketitle

\begin{abstract} 
One of the fundamental problems in machine learning is the estimation
of a probability distribution from data.  Many techniques have been
proposed to study the structure of data, most often building
around the assumption that observations lie on a lower-dimensional
manifold of high probability. It has been more difficult, however, to
exploit this insight to build explicit, tractable density models for
high-dimensional data. In this paper, we introduce the \emph{deep density model}
(DDM), a new approach to density
estimation. We exploit insights from deep learning to construct a
bijective map to a representation space, under which the
transformation of the distribution of the data is approximately
factorized and has identical and known marginal densities. The
simplicity of the latent distribution under the model allows us to
feasibly explore it, and the invertibility of the map to
characterize contraction of measure across it. This
enables us to compute normalized densities for out-of-sample data.
This combination of tractability and flexibility allows us to tackle a
variety of probabilistic tasks on high-dimensional datasets,
including: rapid computation of normalized densities at test-time
without evaluating a partition function; generation of samples without
MCMC; and characterization of
the joint entropy of the data. 
\end{abstract} 

\section{Introduction}
Many core machine learning tasks are concerned with density estimation
and manifold discovery.  Probabilistic graphical models are a
dominating approach for constructing sophisticated density estimates,
but they often present computational difficulties in practice.  For
example, undirected models, such as the Boltzmann
machine~\citep{smolensky1986information, hinton2006fast} are able to
achieve compact and efficiently-computed latent variable
representations at the cost of only providing unnormalized density
estimates.  Directed belief networks~\citep{pearl1988probabilistic,
  neal-1992a, adams-wallach-ghahramani-2010a}, on the other hand,
enable one to specify \emph{a priori} marginals of hidden variables
and are easily normalized, but require costly inference procedures.
Bayesian nonparametric density estimation (e.g.,
\citep{escobar-west-1995a, rasmussen-2000a,
  adams-murray-mackay-2009a}) is another flexible approach, but it
often requires costly inference procedures and does not typically
scale well to high-dimensional data.

Manifold learning provides an alternative way to implicitly
characterize the density of data via a low-dimensional embedding,
e.g., locally-linear embeddding~\citep{roweis2000nonlinear},
IsoMap~\citep{tenenbaum2000global}, the Gaussian process latent
variable model~\citep{lawrence2005probabilistic}, kernel
PCA~\citep{scholkopf1998nonlinear}, and
t-SNE~\citep{vandermaaten2008visualizing}.  Typically, however, these
methods have emphasized visualization as the primary motivation. A
notable exception is the autoencoder neural
network~\citep{cottrell1987autoencoder, hinton-salakhutdinov-2006a},
which seeks embeddings in representation spaces that themselves can be
high dimensional.  Unfortunately, the autoencoder does not have a
clear probabilistic interpretation (although see
\citep{rifai2012sampling} for a discussion).

Some approaches, such as manifold Parzen
windows~\citep{Vincent02manifoldparzen}, have attempted to tackle the
combined problem of density estimation and manifold learning directly,
but have faced difficulties due to the curse of dimensionality.  Other
approaches, such as the Bayesian GP-LVM~\citep{titsias2010bayesian},
characterize the manifold implicitly in terms of a nonlinear mapping
from a representation space to the observed space.  To define a
coherent probabilistic model, however, it is necessary to find an
invertible map between these spaces so that the density of a datum can
be evaluated in the latent space without integrating over the
pre-image.  It has proven difficult to flexibly parameterize the space
of such invertible maps, however, let alone find a transformation that
results in a tractable density on the representation space.
Independent components analysis~\citep{bell1995information} and the
related idea of a density network~\citep{MacKay97densitynetworks} are
examples of bijective models that exploit invertible linear
transformations; these, however, have rather limited expressiveness.
DiffeoMap~\citep{Walder08diffeomorphicdimensionality} establishes a
bijection close to a lower-dimensional subspace, and then projects to
it. Other approaches, such as the the back-constrained
GP-LVM~\citep{lawrence2006backconstrained} attempt to approximate this
bijection.

In this work, we introduce the \emph{deep density model}
(DDM), an approach that bridges manifold discovery and density
estimation. We exploit ideas from deep learning to introduce a rich
and flexible class of bijective transformations of the observed space. 
 We optimize over these transformations
to obtain a map under which the implied distribution on the representation space has an
approximately factorized form with known marginals.  The invertibility of the map
ensures that measure is not collapsed across the transformation, and as such, the
determinant of the Jacobian can be computed. This leads to
fully-normalized probability densities without a partition function.

The combination of rich bijective transformations with density
estimation enables us to explore a variety of modeling directions for
high-dimensional data.  As the approach is generative, we can easily
sample data from a trained model without Markov chain Monte Carlo
(MCMC). We present a variety of applications to the CIFAR
and MNIST datasets, for proof-of-concept. The deep density model also provides new possibilities
for supervised learning by building Bayesian classifiers that have
well-calibrated class-conditional probabilities. This additionally permits to exploit
densities of unlabeled data to perform unsupervised learning, by
constructing mixtures of models and training
them coherently with expectation maximization. 

In developing the deep density model, we also provide
insight into a variety of fundamental concepts for latent variable
models. Using information theoretic tools, we
identify important connections between sparsity
 and the independence of the latent dimensions.
These connections allow finding a map leads to an approximately factorized
latent distribution. By understanding the distribution of the data in
representation space and the transformation that gives rise to it, we
can characterize the entropy of the distribution over data in the
observed space. This enables making informed choices in model
selection. 

\section{Bijections and Normalized Densities}\label{mot}
We are interested in learning a distribution over data in a
high-dimensional space~$\scrY\subseteq\reals^K$.  We denote this
(unknown) distribution as~$\pY(\cdot)$.  An axiomatic assumption in
machine learning is that the data contain structure, and this
corresponds to~$\pY(\cdot)$ distribution having most of its
probability mass on a lower-dimensional, but very complicated,
manifold in~$\scrY$.  Tractably parametrizing the space of such
manifolds and then fitting the resulting distributions to data is a
significant challenge.

Similarly, studying this distribution directly in the observed space presents
both theoretical and computational difficulties. Instead, we consider
how the data might be the result of a transformation from an
unobserved \emph{representation space}~${\scrX\subseteq\reals^K}$.  We
denote the distribution on this space as~$\pX(\cdot)$, and we assume
the observed data arise from a transformation ~${\bbf:\scrX\to\scrY}$.
We assume that the latent distribution~$\pX(\cdot)$ has a
simple factorized form:
\begin{align}
\pX(\rmbx)&=\prod_{k=1}^K \pXk(x_k)\;,
\end{align}
where the marginal factors~$\pXk\cd$ have a simple and known univariate form.

We further make the assumption that~$\bbf\cd$ is bijective and
that~$\bbfi\cd$ is available analytically.  In this case, the
probability density for a point~${\by\in\scrY}$ can be computed:
\begin{align}
  \pY(\rmby) &= \prod_{k=1}^K
  \pXk([\bbfi(\rmby)]_k)\,\left|\frac{\partial\bbf(\rmbx)}{\partial\rmbx}\right|_{\bbfi(\rmby)}. \label{estn}
\end{align}
In this paper, we introduce the \emph{deep density model}, which discovers
rich bijective transformations to map from simple latent
distributions into complex observed densities.  By optimizing over a
large and flexible class of such bijective transformations, it is possible
to discover structure in high-dimensional data sets while still having
a manageable, normalized density estimator.

Bijectivity is critical for ensuring that the density in
Eq.~(\ref{estn}) is normalized.  This bijectivity is in contrast to
many neural network approaches to latent representation where the
latent space is often smaller (for an information bottleneck in, e.g.,
an autoencoder) or larger (for an \emph{overcomplete} representation)
than the observed space.  When the representation space is smaller,
then~$\bbf\cd$ cannot be surjective and so multiple points in~$\scrY$
may map to the same point in~$\scrX$, leading to overestimates of the
density.  In the overcomplete case, we also sacrifice bijectivity
since $\bbfi\cd$ cannot be surjective; in other words, the image of
$\scrY$ under $\bbfi\cd$ will not span $\scrX$. As such, $\pX\cd$ will
have support beyond $\bbfi(\scrY)$.  That is, the latent normalization
includes mass that appears in~$\scrX\setminus\left\{ \bbfi(\scrY)
\right\}$. However, taking this mass into account is a very
challenging problem, whose difficulty unfortunately increases with the
richness of $\bbf\cd$. Thus, by using $\pX\cd$ instead of the
normalized density
\begin{align}
  \frac{ \pX(\rmbx) \mathbb{I}_{\rmbx\in\bbfi(\scrY)} }{\int_{\rmbx\in\bbfi(\scrY)} \pX(\rmbx)\,d\rmbx} < \pX(\rmbx)\;,
\end{align}
we underestimate the true density and are in a situation similar to
that in energy-based models with partition functions.

Thus, when learning a representation of data, we must make a choice
between overcompleteness and bijectivity. Our model will be able to
use bijectivity to produce normalized density estimates, but at the
cost of requiring the representation space to be of equal dimension to
the observed space.  Overcomplete representation spaces are thought to
allow for greater parametric flexibility, but we view the ability to
produce a normalized density estimate as a significant win.

\section{The Deep Density Model}

Our objective is to learn a density estimate on the space~$\scrY$ from
a set of~$N$ training
examples~${\bY:=\{\rmby_{n}\}_{n=1}^{N}\subset\scrY}$.  We seek to
produce an invertible parametric map~${\bbf_{\bTheta}:\scrX\to\scrY}$
such that the observed data under~${\bbfi_{\bTheta}:\scrY\to\scrX}$ are well-described by a simple, factorized
distribution.  We denote this function as the \emph{decoder}, and the
representation-space data as~${\bX:=\{\rmbx_n\}_{n=1}^N\subset\scrX}$. We
construct the map~$\bbf_{\bTheta}$ via a sequence of layers, each of
which applies an invertible linear transformation followed by an
elementwise nonlinearity.  This nonlinearity
must be bijective; we use the standard sigmoid (logistic)
function~${\sigma(z)=1/(1+e^{-z})}$. The parameters for layer~$m$ are
square matrix~${\bOmega_m\in\reals^{K\times K}}$ and bias
vector~${\bomega_m\in\reals^{K}}$.  Layer~$m$ performs
transformation~$\bS_{\bOmega_m,\bomega_m}(\rmby):=\bsigma(\bOmega_m\rmby+\bomega_m)$.
Given~$M$ layers, we write the entire map as the composition
\begin{align}
  \bbf_{\bTheta} := \bigcirc_{m=1}^M\bS_{\bOmega_m,\bomega_m}(\bx)\,,
\end{align}
where~$\bigcirc$ is the composition operator
and~${\bTheta:=\bigcup_{m=1}^M\{\bOmega_m,\bomega_m\}}$.

We also need to construct a non-bijective \emph{encoder}
function~${\bg_{\bPsi}:\scrY\to\scrX}$, which maps the observed data
into the representation space. 
This function will be optimized to imbue the latent distribution~$\pX\cd$
with the properties outlined in Section~\ref{mot}; the necessity of $\bg_{\bPsi}\cd$
will be expanded upon in Subsection \ref{whyg}.
$\bg_{\bPsi}\cd$ is composed of $J$ layers, and the~$j$-th layer of~$\bg_{\bPsi}\cd$ has~$K_j$ hidden
units (with $K_J=K$) and parameters~$\bGamma_j$ and~$\bgamma_j$.  We denote the
parameters for~$\bg$
as~${\bPsi:=\bigcup_{j=1}^J\{\bGamma_j,\bgamma_j\}}$ and use the
notational shortcuts above to write
\begin{align}
  \bg_{\bPsi}(\by) &= \bigcirc_{j=1}^J\bS_{\bGamma_j,\bgamma_j}(\by)\,.
\end{align}

\subsection{Regularizing the Transformations}
We will train the model on data by minimizing an objective composed of
several parts:
\paragraph{Divergence Penalty}
$\mcD(\bPsi)$: This determines the fit of the current encoding
transformation.  It forces the marginal densities of the empirical
distribution of the representation-space data to match a target
distribution of our choice, by penalizing divergence from it.
\paragraph{Invertibility Measure}
$\mcI(\bTheta)$: This ensures the invertibility of~$\bbf_{\bTheta}\cd$ by 
penalizing poorly-conditioned transformations.
\paragraph{Reconstruction Loss}
$\mcR(\bTheta,\bPsi)$: This jointly penalizes the
encoder~$\bg_{\bPsi}\cd$ and decoder~$\bbf_{\bTheta}\cd$ to ensure
that~${\bg_{\bPsi}(\by) \approx\bbfi_{\bTheta}(\by)}$ on the
data. 

Each of these participates in the overall objective given by:
\begin{align}
C(\bTheta,\bPsi) &= \mu_{\mcD}\mcD(\bTheta) +\mu_{\mcI}\mcI(\bPsi)
+\mu_{\mcR}\mcR(\bTheta,\bPsi)\;,\label{obj}
\end{align}
where~$\mu_{\mcI},\mu_{\mcD},\mu_{\mcR}\in\reals$ are the weights of
each term.  We will examine each of these terms in more detail in the
proceeding sections.

\subsection{Divergence Penalty: Sculpting the Latent Marginals}\label{sec:div}
\begin{figure}[t]
\centering
\subfigure[{\small No regularization}]{
\includegraphics[width=2.3cm]{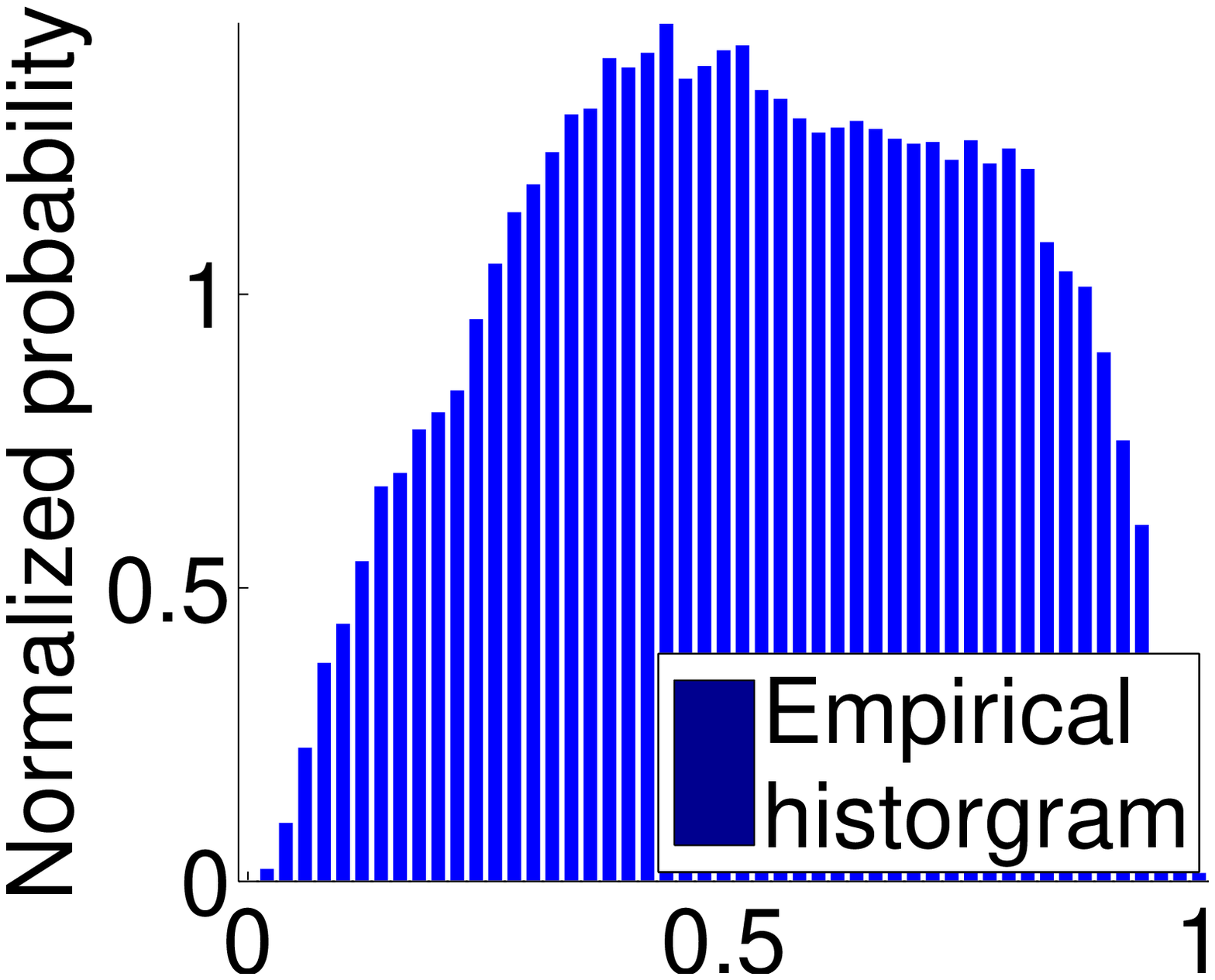}
}\qquad\qquad\qquad
\subfigure[{\small $L_1$ regularization}]{
\includegraphics[width=2.3cm]{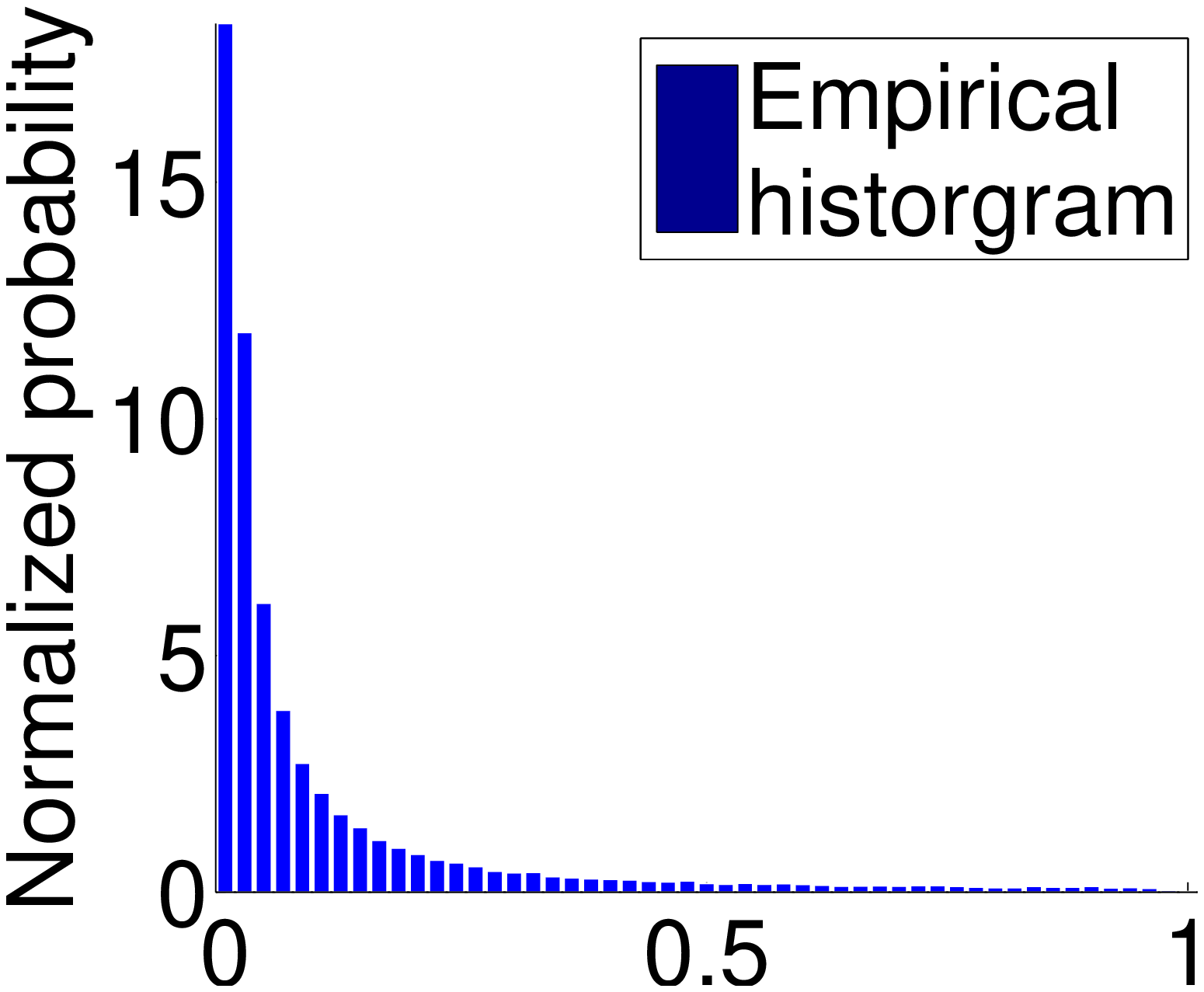} 
}\qquad\qquad\qquad
\subfigure[{\small Divergence penalization}]{
\includegraphics[width=2.3cm]{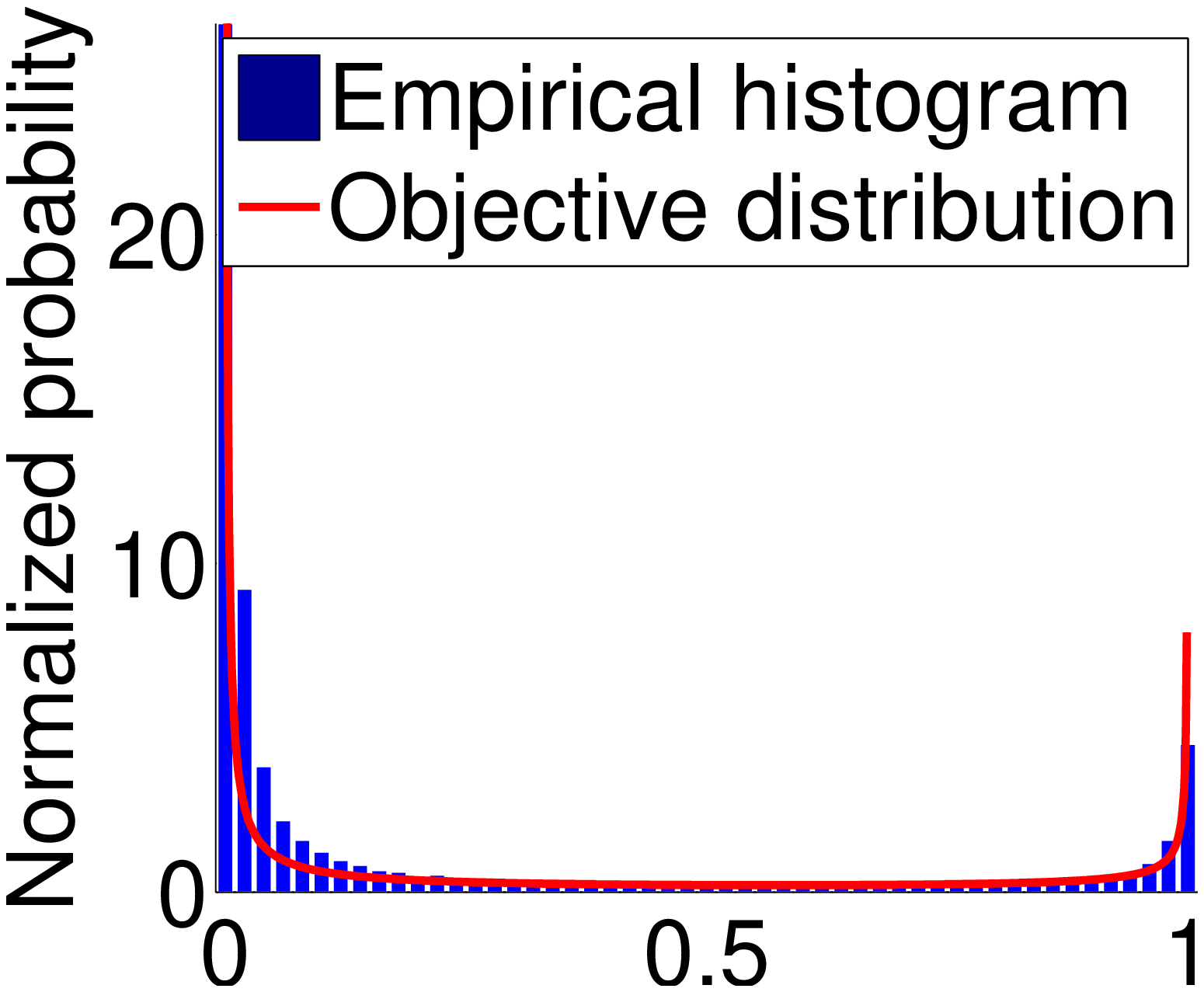}
}
\caption{Histograms of the empirical distribution at $\hpXo\cd,\; k=1$ with the specified methods of regularization, upon training on MNIST. The objective distribution was taken as $\bdist{0.02}{0.2}$.}
\label{regns}
\end{figure}

In order to fit the deep density model to data, it is necessary to
specify a measure of distance between the empirical distribution and
the model distribution.  We achieve this via a divergence penalty,
which forces the model to distribute the mass of the data in the
representation space so as to be similar to a distribution chosen
\emph{a priori}.  This construction has several advantages: it
1)~results in a known, fixed distribution on the representation space
that can be used to generate fantasy data, 2)~enables sparsity to be
enforced as a constraint rather than a penalty weighed against the
reconstruction cost, and 3)~combats overfitting by explicitly
requiring that some of the data have low probability under the
model. This third advantage is subtle, but critical: some data must
live in the tail of the distribution, in contrast to the maximum of
the posterior which is simply a weighing of the MLE against the mode
of the prior. See Figure \ref{regns} for a comparison of 
 distributions produced by a various regularization techniques.

Concretely, we assume that the representation space is a unit
hypercube, i.e., ~${\scrX=[0,1]^K}$.  As before, we assume there
are~$N$ data~$\{\rmby_n\}^N_{n=1}$, which (for a given~$\bPsi$) are
mapped into~$\scrX$ to give~${\{\rmbx_n=\bg_{\bPsi}(\rmby_n)\}^N_{n=1}}$.  Ideally, for
representation dimension $k\in{1,\ldots,K}$, the divergence penalty
would measure the difference between the marginal empirical
distribution
\begin{align}
 \hpXk(x)&= \frac{1}{N}\sum_{n=1}^N\delta(x-[\rmbx_n]_k)
\end{align}
and a target univariate distribution ~$q\cd$ that we define.  In
practice, we approximate~$\hpXk\cd$ by finding the best fit of a
tractable parametric family and then computing the symmetrized
Kullback--Liebler divergence:
\begin{align}
\rdiv{p(\cdot)}{q(\cdot)} &= \div{p(\cdot)}{q(\cdot)} + \div{q(\cdot)}{p(\cdot)}\;,
\end{align}
where
\begin{align}
\div{p(\cdot)}{q(\cdot)} &= \int_{\scrX}p(\rmbx)\,\log\frac{p(\rmbx)}{q(\rmbx)}\,\mathrm{d}\rmbx.
\end{align}
Since, in this case the representation
space is the unit hypercube, we choose our objective distribution to be a member of the Beta family:
\begin{align}
 q(x;\alpha,\beta) &= \frac{\Gamma(\alpha+\beta)}{\Gamma(\alpha)\Gamma(\beta)}
  x^{\alpha-1}(1-x)^{\beta-1}.
\end{align}
Given the data representations~$\{\rmbx_n\}^N_{n=1}$, we estimate the empirical distribution of
each dimension with a Beta distribution, using moment-matching to approximate its parameters:
\begin{align}
\hat{\alpha}_k & =  \hat{\mu}_k\left[\frac{\hat{\mu}_k\left(1-\hat{\mu}_k\right)}{\hat{\sigma}_k^{2}}-1\right]\\
\hat{\beta}_k & =  \left(1-\hat{\mu}_k\right)\left[\frac{\hat{\mu}_k\left(1-\hat{\mu}_k\right)}{\hat{\sigma}_k^{2}}-1\right],
\end{align}
where~$\hat{\mu}_k$ and~$\hat{\sigma}_k^2$ are the sample mean and
variance, respectively. We note that a Beta distribution with parameter $\alpha<1$ produces
a very sharp peak at 0, and as such allows to pursue sparsity in distribution (under the assumption
that elements of small magnitude cannot be distinguished from each other).

With these in hand, we get a closed-form expression for our
divergence penalty:
\begin{eqnarray}
\rdiv{\hpXk(\cdot)}{q(\cdot)} &=&
\left(\hat{\alpha}-\alpha\right)\left[\psi(\hat{\alpha})-\psi(\alpha)\right]+\left(\hat{\beta}-\beta\right)\left[\psi(\hat{\beta})-\psi(\beta)\right]\\
&&-\left(\hat{\alpha}-\alpha+\hat{\beta}-\beta\right)\left[\psi(\hat{\alpha}+\hat{\beta})-\psi(\alpha+\beta)\right],
\end{eqnarray}
where~$\psi(z)=\mathrm{d}\log\Gamma(z)/\mathrm{d}z$ is the digamma
function.

Furthermore, for each example, we impose an example divergence
  penalty, denoted as ~$\hpXn\cd$, which penalizes the distance
between our objective distribution and the empirical distribution over
the elements of that particular example:

Finally, our total divergence penalty is
\begin{align}
\mcD(\bPsi) = \frac{1}{K}\sum_{k=1}^K \rdiv{\hpXk\cd}{q\cd} \nonumber + \frac{1}{N}\sum_{n=1}^N \rdiv{\hpXn\cd}{q\cd}\;.
\end{align}

\subsubsection{Sparsity in Distribution}\label{direct_sp}
The traditional pursuit of sparsity entails the application of an
$L_{1}$-type regularization that directly penalizes the activations in
the representation space. This has several undesirable
properties. First, the penalty does not differentiate between the
cases where the activated
units are distributed evenly among examples, and where a fixed set of units
is always activated at all examples while others never
are. Furthermore, it is
discomforting that, in the limit of small reconstruction cost in the
objective function, the regularization term is optimized if and only
if all examples are identically mapped to the same point, namely
zero. This forces \emph{all} the activations to be small in order to
have some of them vanish; it artificially forces the activation
distribution to be contained in a small region around zero. Another
implication of direct activity penalization is that we must search the
parameter space of the regularization coefficient in order to attain
our desired sparsity structure.

Instead of inducing sparsity directly, we achieve it in
\emph{distribution}, across examples. In practice, this arises from
penalizing the KL-divergence between the \emph{empirical
  distribution}~---~the actual distribution in the representation space
for the given set of observations~---~and an appropriately-chosen
target distribution~$q\cd$, which has a peak at 0.  The difference between this and the
traditional~$L_1$ approach to sparsity can be understood by
considering the optimization problem as the dualization of
the sparsity constraints. In the case of an $L_{1}$-type
regularization, these constraints directly bound the space under the
prescribed distance metric. In the $L_1$ case, we thus have a situation with
two different points on the same contour of these constraints, one
of which a more desirable of a solution than the other. On the other
hand, the constrains that emerge from the divergence penalization are
imposed within the probability simplex. The advantage is that a
contour of these constraints corresponds to a locus of distributions
with similar sparsity structures and thus similar desirability.

\subsection{Invertibility: condition number penalty}

Invertibility of~$\bbf_{\bTheta}$ is critical to providing a
normalized density.  A standard autoencoder contracts volumes around
observed examples only, due to the reconstruction penalty
approximating an invertible map at the observations. A true bijection,
however, will guarantee conservation of volume not just at the data,
but also at points we have never seen before.  This will allow
computation of the determinant of the Jacobian of $\bbf_{\bTheta}\cd$,
and precisely specify how probability mass is reshuffled by the
transformation.  To ensure invertibility of the transformation, we
must ensure the invertibility of each layer.  As the nonlinear
activation functions are fixed, invertibility is determined by the
condition numbers of the~$\bOmega_m$ matrices in~$\bbf_{\bTheta}\cd$.  We
therefore introduce a regularization term that ensures invertibility:
\begin{align}
   \mcI(\bTheta)&=\frac{1}{M} \sum_{m=1}^M \log\left(
   \frac{\lambda_{\max}(\bOmega_m)}{\lambda_{\min}(\bOmega_m)}
   \right).
\end{align}
Here, $\lambda_{\max}(\bA)$ and $\lambda_{\min}(\bA)$ are the maximum
and minimum eigenvalues of $\bA$. In this case, the curse of
dimensionality becomes a blessing of dimensionality: in high
dimensions, not only orthogonality is easily attained, it is difficult
to escape.  We find that in practice the invertibility requirement is
easily satisfied, and does not constrain the algorithm at all.

\subsection{Reconstruction and the independence of latent dimensions}\label{indep}
\begin{figure}[t]
\centering
\subfigure[]{\includegraphics[width=3.5cm]{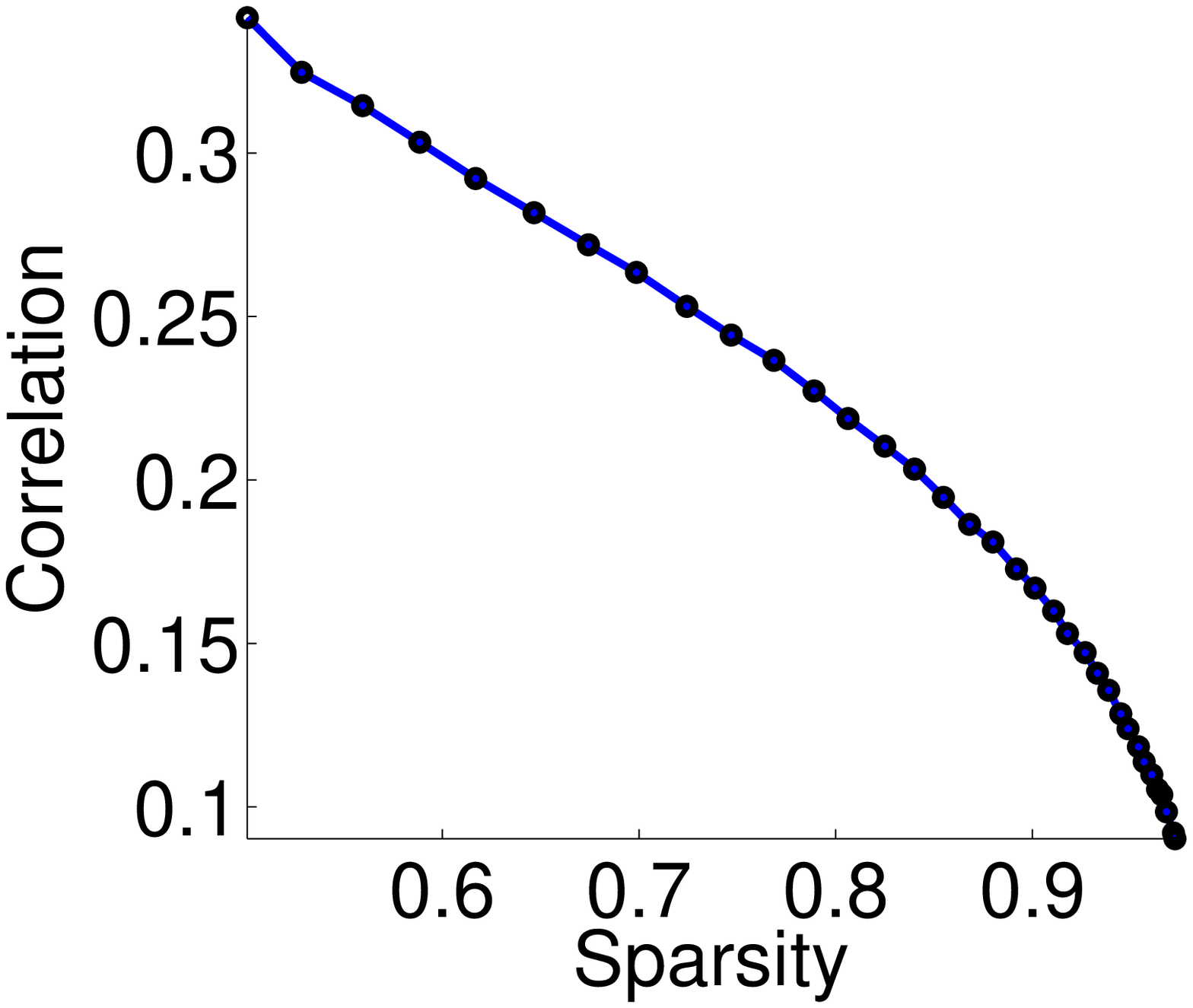}}
\qquad\qquad\qquad
\subfigure[]{\includegraphics[width=4cm]{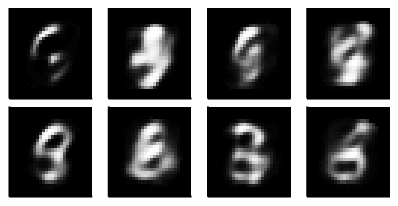}}
\caption{Effects of diversification. (a) Decrease of first-order dependencies as function of sparsity. (b) Generations from the model by sampling independently from the latent marginals. First row: generations from a model with high-entropy marginals $\bdist{0.02}{0.2}$ with strong dependence; images are incoherent. Second row: generations from a model with diversified marginals with a final distribution $\bdist{0.004}{0.2}$. }
\label{diversify}
\end{figure}

Since we now have the ability to dictate the marginal distributions in
the representation space, we can shed light on the connection between
entropy, sparsity and independence as a function of the transformation
to the representation space. In order to attain a tractable
distribution in the representation space, we must eliminate
dependencies between the latent dimensions. We refer to this process
as \emph{diversification}, as it reduces the overlap of information
learnt by distinct dimensions; see the effects of this process in
Figure \ref{diversify}.

Diversification arises naturally when considering the effect of
simultaneously increasing sparsity in the latent representation, while
decreasing the entropy of the target marginals.  The reconstruction
penalty demands that information be preserved in the representation
space.  However, as the entropy decreases, the information capacity of
the marginals decreases; it is necessary for the dimensions of the
latent distribution to become more independent in order to reduce
redundancy and continue to reconstruct successfully.  Hence, we
increase sparsity by limiting the capacity of the encoder, until we
get to the point of minimum marginal entropy under sufficient
conservation of information (which we measure by reconstruction of the
observations). At this point, we expect the marginals to be
approximately independent.

In the second line below, notice that we may write the observed space
entropy in terms of the representation entropy and a term that
accounts for the contraction of volumes under the transformation. We
then write the joint entropy in terms of the marginal entropies and
the mutual information information between the joint distribution and
the independent marginal factorization:
\begin{eqnarray}
\mcH(\pY\cd)&=& -\int_\mcY \pY(\rmby) \log\pY(\rmby) d\rmby\nonumber \\ 
&=&\mcH(\pX\cd)+\bbE_\scrX \left[ \log \left| \prtt{\bbf_{\bTheta}\cd}{\rmby} \right| \right]\nonumber \\
&=&\sum_{k=1}^K \mcH(\pXk\cd) - \div{\prod_{k=1}^K \pXk\cd}{\pX\cd} +\bbE_\scrX \left[ \log \left| \prtt{\bbf_{\bTheta}\cd}{\rmby} \right| \right]  \;.
\end{eqnarray}

The latent dimensions are independent if and only if the mutual information is zero, and as such, we seek to minimize it. The entropy in the observed space is fixed, but we have control of the marginal entropies, and we can decrease ~$\bbE_\scrX\left[ \log \left| \prtt{\bbf\cd}{\rmby} \right| \right]$ by placing an information bottleneck on ~$\bbf\cd$. Thus, by minimizing both these terms, we can minimize latent dependencies. 

To that end, we approximate
\begin{equation}
\bbE_\scrX \left[ \log \left| \prtt{\bbf_{\bTheta}\cd}{\rmby} \right| \right]
\approx \frac{1}{N}\sum_{n=1}^N \log \left| \prtt{\bbf_{\bTheta}(\rmbx_n)}{\rmby} \right|\;.
\end{equation}

Note that an inherent property of the sigmoid nonlinearity is its
asymptotic flatness as its output approaches zero. As such, the
decoder $\bbf_{\bTheta}$ becomes more unable to distinguish between
points as their magnitude decreases. Thus, for each representation
dimension, by increasing the $\alpha$ parameter of our objective
distribution ~$q(\cdot; \alpha,\beta)$, we can shift more probability
mass towards zero: this not only decreases the marginal entropies
$\sum_{k=1}^K \mcH(\pXk\cd)$, but also increases the information
bottleneck. We add that the flatness of the sigmoid still does not
imply true sparsity: a deep transformation can distinguish between
small but nonzero values, and as such "undo" the sparsity
effect---information is not gone, only bit-shifted. As such, in order
to ensure true information loss in the neighbourhood of zero, we also
introduce a threshold to the encoder, that kills any representation
element less than some $\varepsilon>0$; namely, we use the
encoder~${\ind_{\bg_{\bPsi}(\rmby)\geq\varepsilon}
  \bg_{\Psi}(\rmby)}$.

\subsubsection{Entropy characterization}
A direct consequence of the above is our ability to now place an upper bound on ~$\mcH(\pY\cd)$, and, once the mutual
information is minimized, to characterize it. This now allows us to make informed choice of a representation space---for example, we understand the interplay between the latent dimensionality and sparsity. Minimizing the mutual information in the way demonstrated above corresponds to selecting this space to be maximally sparse under the constraint of retaining the information encapsulated in the input examples. 

We furthermore note that the entropy of model can be thought of as the "effective number of configurations" that data drawn from its distribution can take. As such, it governs how mass is allocated between the training examples and out-of-sample data: thus, the diversification procedure described dictates the generalizability of the model in a very transparent way.

\subsubsection{ On why we need $\bg_{\bPsi}\cd$ }\label{whyg}
We can now understand the motivation for introducing~$\bg_{\bPsi}\cd$
even with a bijective map. We need~$\bg_{\bPsi}\cd$ to induce points
in the observed space to be close together in~$\scrX$, both to control
the information capacity as well as to determine the marginals in
$\scrX$. On the other hand,~$\bbf_{\bTheta}\cd$ should not expand
measure in mapping from~$\scrX$ back to~$\scrY$; this ensures that
there is an information bottleneck even under bijectivity.  However,
asking~$\bbf_{\bTheta}\cd$ and~$\bbfi_{\bTheta}\cd$ to serve this
double purpose is contradictory, as one function exactly undoes the
contraction of measure performed by the other
function. Furthermore,~$\bg_{\bPsi}\cd$ is helpful from a
computational point of view. In the process of shaping the latent
distribution, we must define our requirements as penalties on~$\scrX$,
which we proceed to back-propagate. Since asymptotic flatness of
$\bbf_{\bTheta}\cd$ means asymptotic steepness
of~$\bbfi_{\bTheta}\cd$, performing numerical optimization on this
function---let alone demanding representation points to be clustered
in this asymptotically steep regime, as the divergence penalty
requires---is clearly numerically unstable.

\section{Training}

We train the model on a GPU cluster. In cases of continuous input
dimensions, we pre-process the data by whitening and normalizing it.
We additionally use PCA to reduce the dimensionality of CIFAR.

In the pretraining stage, we recursively train single-layer deep
density models with stochastic gradient descent, where we take the
representation space examples of one iteration as the observed space
examples of the next.

In the fine-tuning stage, we optimize the objective in Eq.~(\ref{obj}) by
performing block coordinate descent: we iterate through the layers,
and for each layer we take a step to minimize the objective as a
function only of that layer's parameters.  Since different gradient
magnitudes of distinct layers are not mixed here, this side-steps the
problem of having the gradient exponentially decay during
back-propagation.

The optimization procedure is designed to have few free parameters:
most are actively adapted in the process of optimization. The step
size is chosen adaptively via an inexact Armijo's rule line search.
Exact line search on the overall objective is computationally
expensive and not possible using mini-batches of data.  Secondly, the
coefficients~$\mu_{\mcI},\mu_{\mcD},\mu_{\mcR}$ are adapted to
maintain a specified ratio of gradient magnitudes, as the step
direction is a mixture of the various penalties and is thus dictated
by the relative proportions of their gradients.

We sprinkle masking noise onto the inputs to attain robust solutions,
as presented in \cite{Vincent:2008:ECR:1390156.1390294}. We also add
momentum, but implicitly: we define a window size, and sweep it across
the shuffled indices of the training examples, in increments that
are a fraction of the window size. As such, each example will be presented
in several minibatches in a row, but with each minibatch still introducing
new training examples into the objective. We found this implicit momentum
technique gives the algorithm a more stable convergence.

\subsection{Distribution sequencing and initialization}

Enforcing the divergence penalty immediately after initialization
results in a highly nonconvex and thus a very challenging optimization
problem. The diversification procedure often terminates with an
objective distribution~$\bdist{\alpha}{\beta}$ with~$\alpha\ll1$. As
such, once the algorithm settles near a solution whose empirical
distribution in the representation space takes this shape, it will be
very difficult for it to ``reshuffle'' the representation data to
attain an alternative solution also with the same distribution. To
that end, instead of penalizing directly with the final objective
distribution, we penalize through a family of distributions
$q_{j}\left(\cdot\right)=\bdist{\alpha_{j}}{\beta_{j}}$ over
iterations~$j$, where we have~$\alpha_{0},\beta_{0}>1$ as an ``easy''
initial problem, and then have the parameter sequences converge as
$\alpha_{j}\rightarrow\alpha,\beta_{j}\rightarrow\beta$.  The
transitions between consecutive elements in these sequences are also
adaptive: the objective distribution parameters are updated once the
current empirical distribution is sufficiently close to the objective
distribution.

In addition, we choose the sequence of hyperparameters such that
$\frac{\alpha_{j}}{\alpha_{j}+\beta_{j}}=\frac{\alpha}{\alpha+\beta}=\mathbb{E}\left[\bdist{\alpha}{\beta}\right]$,
as maintaining a constant expectation improves stability.  In
accordance with this, we initialize the biases as
$\log\left(\frac{\alpha}{\beta}\right)$, which similarly gives rise to
a consistent initial expectation. In order to initialize the weight
matrices to satisfy the invertibility constraint, we select them to be
random orthonormal matrices, with a scaling such that the
representation distribution approximately matches in variance the
initial objective distribution~$\bdist{\alpha_{0}}{\beta_{0}}$.

\section{Empirical results}

In this section, we examine the properties of the model and learning
algorithm on benchmark data: the MNIST digits (60,000 ${28\times 28}$
binary handwritten
numerals)\footnote{\url{http://yann.lecun.com/exdb/mnist/}} and the
CIFAR-10 image
data\footnote{\url{http://www.cs.toronto.edu/~kriz/cifar.html}}
(50,000 ${32\times 32}$ color images).  In particular, we examine the
global and local properties of the learned densities, via generation
and perturbation.  We emphasize that the density estimates we report
here arise are fully normalized due to the tractability of the
Jacobian determinant of our transformation.

\begin{table}[t]
\begin{centering}
\begin{tabular}{|c||c|c|}
\hline 
Dataset & MNIST & CIFAR-10 \tabularnewline
\hline
\hline
Training examples & $3301.5$ & -41.8  \tabularnewline
\hline
Test examples (TE) & $3343.7$ &	-45.5  \tabularnewline
\hline
\tablebreak{Test examples \\ rotated by $90^{0}$} & $\mbox{(TE)}-63.9$ & $\mbox{(TE)}-1.9$		\tabularnewline
\hline
\tablebreak{TE with 10\% of \\ elements corrupted} & $\mbox{(TE)}-310.6$ & $\mbox{(TE)}-123.1$		\tabularnewline
\hline
\end{tabular}
\par\end{centering}
\caption{Mean log-probabilities of points in the observed space. The test examples are assigned similar probability as the training examples; the rotated test examples are assigned slightly less probability than the test examples; and the corrupted examples are assigned significantly lower probability. Both models were taken to have 3 layers.}
\label{distortions}
\end{table}

\subsection{Density evaluation}

We start by conducting a variety of tests to examine the quality of
the density estimates produced by the DDM.

First, for a probability model to be useful, it should not overfit by
distributing most of the mass to training examples, but also assign
high probability to unseen out-of-sample data.  Similarly, it should
assign low density to points in data space that resemble real
observations, which in fact are not.

In Table \ref{distortions}, we compute the probabilities assigned by
models to their training examples, test examples, and distortions of
the test examples.

\begin{figure}[t]
\centering
\subfigure[]{\includegraphics[width=3.5cm]{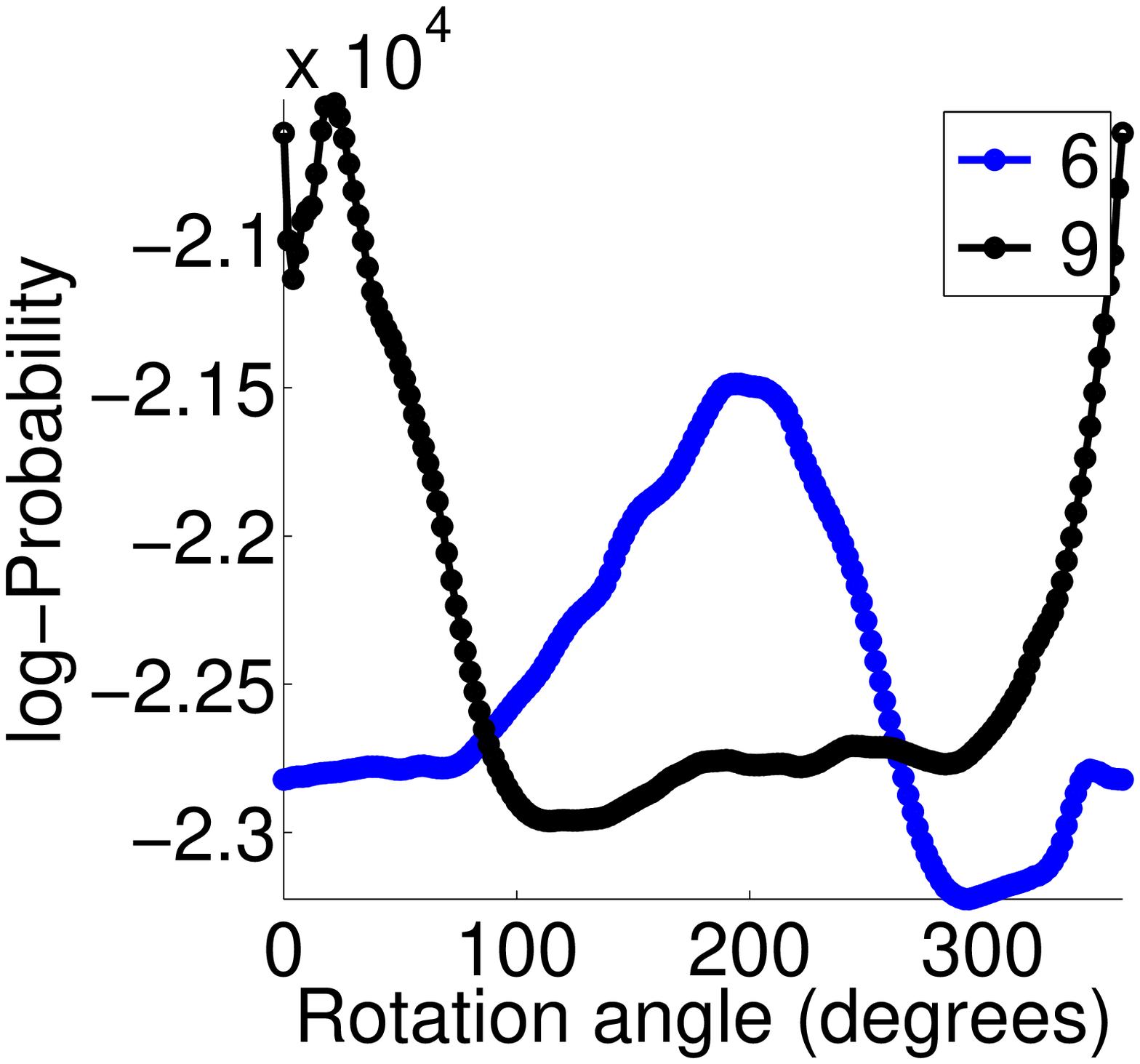}}
\qquad\qquad\qquad
\subfigure[]{\includegraphics[width=3cm]{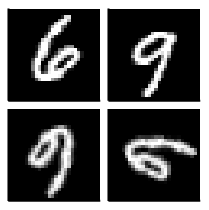}}
\caption{Probabilities of a rotated 6 under a model trained only on 9's. A real 9 is more probable than an inverted 6. (a) log-densities of a 6 and a 9 as a function of the angle of rotation. (b) Respectively: the original 6; a 9; the rotation of the 6 that the 9-model assigned the highest probability to; and the rotation that the 9-model assigned the lowest probability to.}
\label{rot}
\end{figure}

As an interpretable example, we train a model on examples from MNIST's
digit 9 class, and consider the log-probability it assigns the digit 6
under rotation. Intuitively, we expect the highest density to be
assigned to the upside-down 6; see Figure \ref{rot}. We also add a
test-set 9 to demonstrate the density calibration.

We also investigate the marginal entropy of MNIST.  We train the model
to have a final diversification marginals $\bdist{0.01}{0.4}$.  This
distribution is extremely peaked at 0 and 1, which justifies
approximating the MNIST representation elements as Bernoullis by
rounding off to 0 and 1 as $\left[x_{n}\right]_{k}=\left\lceil
\left[x_{n}\right]_{k}-\frac{1}{2}\right\rceil $, which corresponds to
Bernoulli parameter
$p=\int_{1/2}^{1}\frac{\Gamma\left(0.01+0.4\right)}{\Gamma\left(0.01\right)\Gamma\left(0.4\right)}x^{0.01-1}\left(1-x\right)^{0.4-1}d\mbox{x}\approx0.0224$.
We now note that, by the law of large numbers, as
$K\rightarrow\infty$, we expect the mean log-probability of the
Bernoulli test examples to approach
$-KH\left(\mbox{Bern}\left(\cdot;0.0465\right)\right)=21.0181...$.
Indeed, we compute that
$\frac{1}{N}\sum_{n=1}^{N}\sum_{k=1}^{K}\log\mbox{Bern}\left(\left[x_{n}\right]_{k};0.0465\right)=20.7217...$.
Thus, we see that the marginal entropy of the model is close to our
expectation of it.

\subsection{Generation}

\begin{figure}[h]
\centering
\subfigure[Generations from a 3-layer DDM on MNIST, with diversified marginals $\bdist{0.015}{0.8}$.]{
\includegraphics[width=8cm]{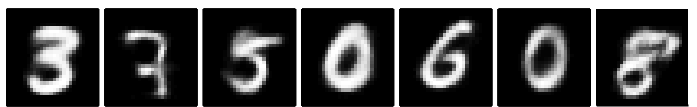}}\\
\subfigure[Generations from a 3-layer DDM on CIFAR-10, with diversified marginals $\bdist{0.5}{3}$.]{
\includegraphics[width=8cm]{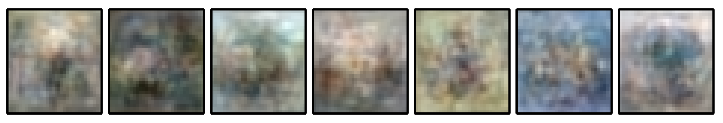}}
\caption{Generations from DDMs.}
\label{generations}
\end{figure}

The approximate independence due to the process of diversification described in Subsection \ref{indep} combined with
the invertibility property of the decoder allows us to produce real samples from the
density estimation on the observed space, by sampling within the representation space. Such instantiations of the
distribution provide us with visualizations of regions of high density under it; see Figure \ref{generations}.

\section{Discussion}

In this paper, we have proposed a new way to construct normalized probability
density estimates from high-dimensional data.  Our approach borrows
ideas from deep learning, differential geometry, and information
theory to ensure that the learned distributions are rich, but
tractable and normalizable.  

There are several interesting directions
that we believe such a density model opens up. One advantage of a fully-normalized density model is that it enables
Bayesian probabilistic classifiers to be constructed to provide
calibrated conditional distributions over class membership.  If the
data are drawn from a mixture of $C$ classes, we may compute
$p\left(\cdot;c\right)$, the likelihood that an element is drawn from
class $c$ by training a model with only the examples under that
class. It is then possible to, e.g., estimate the most likely class:
\[
c_{n}^{*}=\arg\max_{c}p\left(\rmby_n;c\right)\qquad\forall n=1,\ldots,N\;.
\]
Moreover, since we now have class-conditional densities, we may reject
classification of examples if no class model is confident enough to
own them.  Namely, we can set a threshold parameter $\lambda$, and
reject classification of example $\rmby_n$ if
$p\left(\rmby_n;c\right)<\lambda\;\forall c=1,\ldots,C$.

Lastly, we can not only control the density at the observations, but
also the distribution of mass in the rest of the latent space.
Specifically, instead of letting each class model see only its own
examples, we can expose it to training examples from other classes,
and demand it assigns them as little probability as possible. This not
only teaches the model to recognize examples that lie on the manifold
of its class, but also identify differences from other examples by
mapping them away from this manifold.

We tested the above ideas on MNIST, and found that a raw mixture of
Bayesian classifiers gives us an error rate of 9.5\%. However,
penalizing density assigned to foreign examples results in a model
that achieves a much lower error rate of 1.614\%.  As the model
provides calibrated probabilistic predictions, it is also able to
assess its confidence when making classifications.  Among examples in
which the model is confident (approximately 95\% of the test data),
the Bayesian DDM classifier achieves 0.45\% error.  Although these are
not state-of-the-art rates, they show the flexibility of classifier
construction in the DDM setting and how the normalized density can be
leveraged \emph{across} separately trained models, something not
typically possible for energy-based probabilistic approaches.

We can extend these ideas further to use the deep density model even
if only a small fraction of the observations are labelled.  That is,
we can use density estimates to extract useful information from
unlabeled data by leveraging our knowledge of the empirical density
in the representation space.  One possible approach is to run the
expectation-maximization algorithm and train the DDM on weighted data
as part of a mixture model.

\bibliography{deep_density_models_arxiv}
\bibliographystyle{unsrtnat}

\end{document}